\documentclass[runningheads]{llncs}
\usepackage{cite}
\usepackage{amsmath,amssymb,amsfonts}
\usepackage{algorithmic}
\usepackage{graphicx}
\usepackage{pdfpages}
\usepackage{epstopdf}
\usepackage{float}
\usepackage{textcomp}
\usepackage{xcolor}
\usepackage{booktabs}
\usepackage{multirow}
\usepackage[normalem]{ulem}
\useunder{\uline}{\ul}{}
\usepackage{subfigure} 

\begin{document}
\title{A high-order tensor completion algorithm based on Fully-Connected Tensor Network weighted optimization\thanks{Correspondence author. E-mail address: $msh\_huizi@gdut.edu.cn$}}
\author{Peilin Yang \and
Yonghui Huang\inst{*} \and
Yuning Qiu \and
Weijun Sun \and
Guoxu Zhou}
\authorrunning{~~}
\institute{School of Automation, Guangdong University of Technology, Guangzhou 510006, China}
\maketitle              % typeset the header of the contribution

\begin{abstract}
    Tensor completion aimes at recovering missing data, and it is one of the popular concerns in deep learning
    and signal processing. Among the higher-order tensor decomposition algorithms, the recently proposed fully-connected tensor network 
    decomposition (FCTN) algorithm is the most advanced. 
    In this paper, by leveraging the superior expression of the fully-connected tensor network (FCTN) decomposition, we propose a new tensor completion method 
    named the fully connected tensor network weighted optization(FCTN-WOPT). 
    The algorithm performs a composition of the completed tensor by initialising the factors from the FCTN decomposition. We build a loss function with the weight tensor, the completed 
    tensor and the incomplete tensor together, and then update the completed tensor using the lbfgs gradient descent algorithm  to reduce the 
    spatial memory occupation and speed up iterations. 
    Finally we test the completion with synthetic data and real data (both image data and video data) and the 
    results show the advanced performance of our FCTN-WOPT when it is applied to higher-order tensor completion.
\end{abstract}
\section{Introduction}
Higher-order tensor completion is the prediction of missing data from the original tensor. With 
the development of Internet technology and artificial intelligence, higher-order data is gradually 
spreading throughout the various fields of scientific research and engineering applications. In reality, 
for example, a video (length $\times$ width $\times$ number of frames) or a colour picture (length $\times$ width $\times$ number of channels) 
is a third-order data, also known as a third-order tensor.
When the data is high-dimensional, it can be compressed by tensor decomposition algorithms to reduce 
the space occupation, and tensor completion is one of the applications of tensor decomposition. Tensor has been studied for more than a century and 
is also widely used in engineering for neural networks \cite{b8}, machine learning \cite{b10}, computer 
vision \cite{b5}, biosignal processing \cite{b9}, image processing \cite{b6}, etc.

An early proposal is the tensor CANDECOMP/PARAFAC (CP) decomposition \cite{b1}, which decomposes a tensor into a sum of rank-one 
tensors. It is able to represent a large amount of data with a small amount of data. However, the drawbacks of CP decomposition have gradually become apparent when finding the optimal latent factors is very difficult. In recent 
years, the TT decomposition \cite{b2} algorithm has been proposed for higher order tensors (greater than or equal to third order), which can decompose (n-2) 
third order tensors and 2 matrices and concatenate them into the shape of a train. Currently, a generalization of TT decomposition, termed the
tensor ring (TR) decomposition\cite{b3}, has been studied across scientific disciplines. This method of decomposition has also been applied to different areas\cite{b20}\cite{b21}\cite{b22}. 
Zheng et al. also proposed the FCTN fully connected decomposition algorithm \cite{b11}, which factors are interconnected with each other forming a net. 
Their algorithm guarantees the potential correlation of all factors and has one of the best performances in tensor completion.

Tensor completion is one of the important applications of tensor decomposition, where the goal is to recover incomplete tensors from the observed data. Tensor completion can become quite challenging when observations are few and scattered. 
Tensor recovery algorithms on this highly undersampled tensor present additional computational and theoretical challenges, see \cite{b4}.
The key theories of tensor completion currently fall into two main categories: the first is an algorithm based on nuclear norm approximation, the other is based on a low-rank tensor decomposition.
\subsection{Nuclear Norm Approximation:}

Over the past decade, the results of a large number of research experiments have confirmed that a low-rank tensor can substantially improve the effectiveness of 
tensor completion. We use $\mathcal{X}$ to denote the recovered low-rank tensor, $\mathcal{T}$ to denote the observed tensor, $\Omega$ to 
denote the location coordinates of the observed tensor data, and $P_\Omega(~)$ to denote the observed data.
The classical low-rank tensor completion (LRTC) model is mathematically represented as:
\begin{equation}
    \begin{aligned}
    & \min\limits_{\mathcal{X}} \quad rank(\mathcal{X})\\
    & s.t. P_\Omega(\mathcal{X}) = P_\Omega(\mathcal{T})
    \end{aligned}
\end{equation}

This approach is a good idea, but the problem is non-convex and NP-hard when solving it computationally, and many difficulties are encountered when solving it. The researchers 
have therefore proposed a tensor nuclear norm to approximate the rank of the tensor, so that the original problem (1) can be equated as
\begin{equation}
    \begin{aligned}
    & \min\limits_{\mathcal{X}} \quad \sum\limits^N_{i=1}\|{\bf{X}}_{(i)}\|_*\\
    & s.t. P_\Omega(\mathcal{X}) = P_\Omega(\mathcal{T})
    \end{aligned}
\end{equation}
where $\|\quad\|_*$ denotes the tensor nuclear norm operation, generally defined as the sum of the singular values of the matrix ${\bf{X}}_{(i)}$. This model is 
the convex problem \cite{b7}, which is easy to solve. This type of model laid the foundation for tensor nuclear norm, and many related algorithms were proposed to follow.
For example, Hu et al. proposed a twisted tensor nuclear norm completion \cite{b12}, which unfolds the tensor in two dimensions to calculate its nuclear norm for 
two-dimensional matrices; Yuan et al. proposed to unfold the tensor under different mode directions \cite{b13} and calculate their nuclear norm separately; 
Yu et al. proposed a TR-rank nuclear norm unfolding \cite{b14}, which has good performance in terms of the effectiveness.

\subsubsection{Low-rank tensor decomposition} 

Tensor decomposition-based algorithms do not apply a low-rank constraint on the target tensor; Instead, they decompose the observed incomplete tensor and the small 
tensor obtained from the decomposition is reconstructed to predict the original tensor. A more classical model framework for tensor complementation algorithms based on decomposition 
is as follows:
\begin{equation}
    \begin{aligned}
        \mathop {\min }\limits_{\mathcal{G}^{(1)},\mathcal{G}^{(2)},\dots,\mathcal{G}^{(N)}} & \quad \|\mathcal{T}-\mathcal{X} (\mathcal{G}^{(1)} , \mathcal{G}^{(2)}, \dots,\mathcal{G}^{(n)})\|^2_F\\
    s.t. & ~~~~~~~P_\Omega(\mathcal{X}) = P_\Omega(\mathcal{T})
    \end{aligned}
\end{equation}
where $\mathcal{G}^{(n)}$ within Eq. denotes the $n$th factor obtained by tensor decomposition, which can be obtained by some synthetic algorithm to obtain the completed 
tensor $\mathcal{X}$ and each $\mathcal{G}$ is a tensor decomposed by the same decomposition method. Based on different decomposition models, different completion algorithms 
are obtained. Such as CP weighted optimization, Tucker weighted optimization, etc. Based on TR decomposition Zhao et al. also proposed the TRALS algorithm using alternating 
lowermost squares (ALS) for iterative optimization \cite{b3}, followed by Yuan et al. \cite{b15} who proposed the TRWOPT algorithm using the gradient descent algorithm for 
solution, all of which achieved good complementary results. Recently, Liu et al. \cite{b16} proposed a robust completion method that can separate the noise from the original 
tensor; Ahad et al. \cite{b17} proposed a residual constraint to improve the model for completion, which also achieved good results.

However, for the factor decomposed by these models, the correlation between them is weak, resulting in more data needing to be decomposed to recover the predicted accurate 
tensor. The recent proposed tensor fully connected decomposition (FCTN) algorithm \cite{b11} solves this problem well. The algorithm has better data correlation and 
compressibility. However, the FCTN algorithm is very harsh in terms of the constraints on tensor completion, and we hope to construct the model with some weaker constraints, 
which may be able to better satisfy the low rank of the tensor and get some better experimental results. 
Therefore, we propose the FCTN-WOPT algorithm to optimise tensor completion and provide better completion results for experiments.

Our proposed algorithm has good recovery results in a variety of experimental situations, and the main innovations and contributions of this paper are:
\begin{itemize}
    \item The FCTN-WOPT algorithm proposed in this paper extends the recently proposed FCTN algorithm to the field of tensor complementation and obtains better optimisation results.
    \item This paper uses a gradient descent algorithm with a reduced spatial memory occupation for accelerated iterations to speed up the algorithm.
    \item This paper experimentally verifies that the model is able to iterate to its theoretical optimum using gradient descent.
\end{itemize}

\section{PRELIMINARIES}
\subsection{Notations}

In this article, we use $x$, $\mathbf{x}$, $\bf{X}$ and $\mathcal{X}$ to denote scalars, vectors, matrices and tensors respectively. Similarly, a Nth-order tensor, we denote 
by $\mathcal{X} \in R^{I_1 \times I_2 \times \dots \times I_N}$. The $(i_1,i_2,\dots,i_N)$ elements of the tensor $\mathcal{X}$ are denoted as $\mathcal{X}(i_1,i_2,\dots,i_N)$.
$\mathcal{X}*\mathcal{Y}$ denotes the Hadamard product of tensors of the same size $\mathcal{X}$ and $\mathcal{Y}$. $\|\mathcal{X}\|_F=\sqrt {\sum_{i_1,i_2,\dots,i_N} |\mathcal{X}(i_1,i_2,\dots,i_N)|^2} $ 
denotes the Frobenius norm of the tensor $\mathcal{X}$. The k-mode of the tensor $\mathcal{X}$ unfolds as a matrix, written as ${\bf{X}}_{(k)}\in R^{k \times I_1 I_2\dots I_{k-1} I_{k+1}\dots I_N}$.

\subsection{Fully-Connected Tensor Network Decomposition}

The FCTN decomposition \cite{b11}, recently proposed by Zheg et al. is an improvement on the TR decomposition. Their algorithm strengthens the connection between the factor 
and better compresses the amount of model data. The purpose of the FCTN decomposition is to decompose the nth-order tensor $\mathcal{X} \in R^{I_1 \times \dots \times I_N}$ 
into a set of $N$th-order factors $\mathcal{G}^{(k)} \in R^{R_{1,k} \times \cdots R_{k-1,k} \times I_k \times R_{k,k+1} \times \cdots \times R_{k,N}}$, which we 
specify in the following form:
\begin{equation}
    \begin{aligned}
        & \mathcal{X}(i_1,i_2,\dots,I_N) = \\
        & \sum\limits^{R_{1,2}}_{r_{1,2=1}}\sum\limits^{R_{1,3}}_{r_{1,2=1}} \dots 
        \sum\limits^{R_{1,N}}_{r_{1,N=1}} \sum\limits^{R_{2,3}}_{r_{2,3=1}}
        \dots \sum\limits^{R_{2,N}}_{r_{2,N=1}}\dots \sum\limits^{R_{N-1,N}}_{r_{N-1,N=1}}\\
        & \{\mathcal{G}^{(1)}(i_1,r_{1,2},r_{1,3},\dots,r_{1,N})\\
        & \mathcal{G}^{(2)}(i_{1,2},r_2,r_{2,3},\dots,r_{2,N})\dots \\
        & \mathcal{G}^{(k)}(r_{1,k},r_{2,k},\dots,r_{k-1,k},i_k,r_{k,k+1},\dots,r_{k,N})\dots\\
        & \mathcal{G}^{(n)}(r_{1,N},r_{2,N},\dots,r_{N-1,N},i_N)\}. 
    \end{aligned}
\end{equation}

To introduce the FCTN decomposition \cite{b11} more dynamically, we use Figure 1 to carry out a representation of the FCTN decomposition form of an Nth-order tensor. This 
decomposition can be seen to be an ordinary matrix decomposition when the decomposition tensor is a matrix. When the decomposed tensor is a third-order tensor, it will 
be seen that the decomposition form is that of the standard TR decomposition \cite{b3}.

\begin{figure}
    \centering
    \includegraphics[width=8cm]{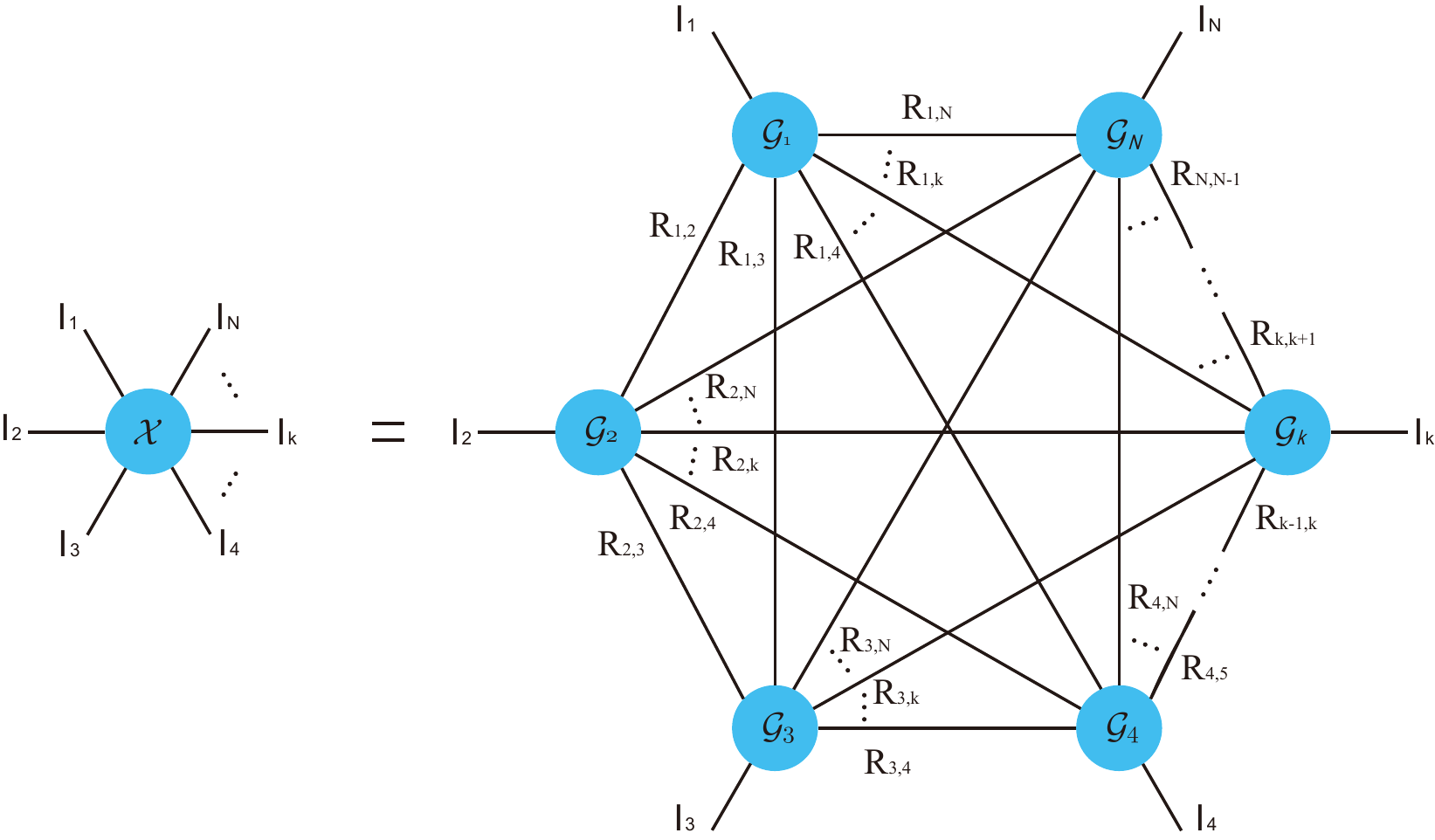}
    \caption{FCTN decomposition}
    \label{1}
\end{figure}

\section{Fully connected tensor network weighted optization}
For the tensor completion problem, based on the FCTN decomposition, we propose a  FCTN weighted optization algorithm. The specific model is as follows:
\begin{equation}
    \mathop {\min }\limits_{\mathcal{G}^{(1)},\mathcal{G}^{(2)},\dots,\mathcal{G}^{(N)}} \quad \|\mathcal{W}*(\mathcal{T}-FCTN(\{\mathcal{G}^{(n)}\}^N_{n=1} ))\|^2_F
\end{equation}
where $\mathcal{T}$ is the tensor of observed partial data and $\mathcal{W}$ denotes the position tensor of observable data with individual elements of 0 or 1.
$FCTN(\{\mathcal{G}^{(n)}\}^N_{n=1})$ is an approximate tensor which is composed by the FCTN decomposition algorithm, and $\{(\mathcal{G}^{(n)})^N_{n=1}\}$ 
is expressed as a total of $N$ factor. And we record the formula more concisely by using $\mathcal{X} = FCTN(\{\mathcal{G}^{(n)}\}^N_{n=1})$.

According to Definition (5) of the original FCTN paper, if one of the factors $\mathcal{G}^{(t)}(t\in \{1,2,\dots ,N\})$, does not participate in the composition, we
denote it as $FCTN(\{\mathcal{G}^{(n)}\}^N_{n=1}, \mathcal{G}^{(t)})$. And we record the formula more concisely by using $\mathcal{M}_t = FCTN(\{\mathcal{G}^{(n)}\}^N_{n=1}, \mathcal{G}^{(t)})$.
The t-mode unfolding of $\mathcal{X}$ he can be written as the following equation.

\begin{equation}
    {{\bf{X}}_{(t)}} = {\left( {{{\bf{G}}_t}} \right)_{(t)}}{\left( {{{\bf{M}}_t}} \right)_{\left[ {{m_{1:N - 1}};{n_{1:N - 1}}} \right]}}
\end{equation}

\noindent where
\[{m_i} = \left\{ {\begin{array}{*{20}{l}}
    {2i,}&{{\rm{ if }}~i < t,}\\
    {2i - 1,}&{{\rm{ if }}~i \ge t,}
    \end{array}} \right.{\rm{  }}and~{\rm{  }}{n_i} = \left\{ {\begin{array}{*{20}{l}}
    {2i - 1,}&{{\rm{ if }}~i < t}\\
    {2i,}&{{\rm{ if }}~i \ge t}
    \end{array}} \right.\]
where $(G_t)_{(t)}$ denotes the t-mode unfolding of the $t$th tensor $\mathcal{G}$ and ${\left( {{{\bf{M}}_t}} \right)_{\left[ {{m_{1:N - 1}};{n_{1:N - 1}}} \right]}}$ denotes a special 
unfolding of the tensor $\mathcal{M}$ according to the above equation, which is abbreviated to $({\bf{M}}_t)_{(\neq t)}$ for simplicity in recording it. This unfolding of the equation is 
very important and the algorithm needs to use it in the derivative calculations.

Since each $\mathcal{G}$ is independent, we can optimize each $\mathcal{G}$ individually, and equation (5) is optimized as
\begin{equation}
    \begin{aligned}
        \mathop {\min }\limits_{({\bf{G}}_t)_{(t)}} ~f\left( {({{\bf{G}}_t})_{(t)}} \right) = \frac{1}{2}\parallel {\bf{W}}_{(t)}*({{\bf{T}}_{(t)}} - {\left( {{{\bf{G}}_t}} \right)_{(t)}}{\left( {{{\bf{M}}_t}} \right)_{(\neq t)}})\parallel _F^2
    \end{aligned}
\end{equation}

Similar to that of Eq. (7), we can obtain the equations for the other factor $\mathcal{G}$. We then take the partial derivative of Eq. (7) for each factor tensor:
\begin{equation}
    \begin{aligned}
        \frac{{\partial f}}{{\partial {{\left( {{{\bf{G}}_t}} \right)}_{(t)}}}} = \left( {{{\bf{W}}_{(t)}}*\left( {{{\left( {{{\bf{G}}_t}} \right)}_{(t)}}{{\left( {{{\bf{M}}_t}} \right)}_{(\neq t)}} - {{\bf{T}}_{(t)}}} \right){{\left( {{{\bf{M}}_t}} \right)^T}_{(\neq t)}}} \right).
    \end{aligned}
\end{equation}

After finding the gradient of the above equation for each $t = 1,2,\dots,N$, we can iterate over the original equation (5) with gradient descent and exit the iterative program when the 
iteration termination condition is satisfied, returning to obtain our recovery tensor. When the sampling rate is 1, the final result achieved using this algorithm is the result of the 
FCTN\cite{b11} decomposition. We give the algorithm flowchart1 below, the specific iteration parameter settings for using gradient descent and some experimental details of the paper 
will be explained in the experimental section.

\begin{table}[!htbp]
    \centering
    \begin{tabular}{l}
    \toprule[1.5pt]
    \textbf{Algorithm 1} LBFGS-Based Solver for FCTN-WOPT\\
    \midrule[1pt]
     1: ~~ \textbf{Input}:The observed tensor $\mathcal{T}$, the weighted tensor $\mathcal{W}$\\
    ~~~~~~~, the matrix $\bf{R}$ and the termination conditions $\bf{opt}$\\
    2:  ~~\textbf{While} Satisfying the termination conditions $\bf{opt}$\\
    3:  ~~~~ \textbf{For} n=1:N \\
    4:   ~~~~~~~~Calculate gradients of each $(G_t)_{(t)}$ using equation (8)\\
    5:  ~~~~ \textbf{End}\\
    6: ~~Update each $\mathcal{G}$ by lbfgs gradient descent\\
    7: ~~\textbf{End while}\\
    8:~~$\mathcal{X} = {P_\Omega }(\mathcal{T} ) + {P_{\bar \Omega }}(FCTN(\{\mathcal{G}^{(n)}\}^N_{n=1} ))$\\
    9: ~~\textbf{Output}:Recovered tensor$\mathcal{X}$\\
    \bottomrule[1.5pt]
    \end{tabular}
\end{table}

\section{Experimental}

We tested tensor completion using different data: synthetic data, image data and video data. For the 
synthetic data, we tested the tensor at different orders. The real data then includes the image data 
and the video data. The termination conditions are set as follows: maxiter = 200 and tol = $10^{-5}$. 
where the tol parameter we define as follows.
\begin{equation}
    tol = \|\mathcal{T}-\mathcal{X}\|_F/\|\mathcal{X}\|_F
\end{equation}
where $\mathcal{T}$ denotes the real data tensor and $\mathcal{X}$ denotes the recovery tensor.

Two parameters were chosen to evaluate the experimental results: the Peak Signal to Noise Ratio (PSNR) 
and the Structural Similarity (SSIM) evaluation criteria.
Where PSNR is defined as follows:
\begin{equation}
    PSNR = 10log_{10}(255^2/MSE)
\end{equation}
MSE is defined as follows:
\begin{equation}
    MSE = \|\mathcal{T}-\mathcal{X}\|_F/num(\mathcal{T})
\end{equation}
where num() denotes the number of tensor elements. We sample the original tensor at random, at a 
sampling rate set by ourselves. The higher the PSNR value, the higher the image recovery quality. 
Also, the definition of SSIM is given:
\begin{equation}
    SSIM(x,y) = \frac{{(2\mu_x\mu_y+c_1)(2\sigma_{xy}+c_2)}}{{(\mu^2_x+\mu^2_y+c_1)(\sigma_x^2+\sigma_y^2+c_2)}}
\end{equation}
where $\mu_x$ is the mean of $x$, $\mu_y$ is the mean of $y$, $\sigma^2_x$ is the variance 
of $x$, $\sigma^2_y$ is the variance of $y$, and $\sigma_{xy}$ is the $x$ and $y$ of the covariance. 
$c_1=(K_1L)^2$, $c_2=(K_2L)^2$ are the constants used to keep the stability.$L$ is the dynamic range of the
 pixel values. $k_1 = 0.01$ and $k_2 = 0.03$. This parameter indicates the similarity 
 between the two tensors and has a value between -1 and 1. When the tensor are the same, the value of SSIM is 1.

 The experimental algorithms we have chosen to compare are the CP-WOPT\cite{b18}, TR-WOPT, TRALS and TRLRF\cite{b19} 
 algorithms, which are all approximate to the methods proposed in this paper and currently have a good algorithmic 
 result. Among them, the CP-WOPT and TR-WOPT algorithms are weighted completion methods based on the 
 CP and TR decompositions. TRALS was an optimised completion algorithm when Zhao et al. proposed the 
 TR decomposition, however these rank choices are difficult and time consuming, later Yuan et al. proposed the 
 TRLRF algorithm for completion, their algorithm makes a low rank constraint on each TR decomposition out of 
 the factors and automatically finds a low rank factor tensor that possesses a better completion outcome. All 
 these algorithms are well referenced and advanced, and we use them for comparison and reference.

\subsection{Synthetic data experiments}
We experimented with this data using classic lena images manipulated by manual reshape into tensors of various 
orders, which we tested at different orders: 120 × 120 × 21 (3-D), 60 × 60 × 20 × 20 (4-D), 
20 × 20 × 5 × 5 × 5 × 5 (5-D) and 5 × 5 × 3 × 3 × 3 × 3 × 3 (6-D). We used the sampling rate as the horizontal axis, 
ranging from 0.1 to 0.9 with an interval of 0.1, and used the PSNR as the vertical axis for presentation. For 
all the algorithms compared, we use the same rank, R is set to 3, and the tensor rank matrix of our proposed 
algorithm is also set to 3. Both maxiter and tol are the same, set to 500 and $10^{-4}$. Different coloured 
lines are used to denote different methods, and we test these five algorithms as shown in Figure 2.

\begin{figure}[H]
    \centering
    \subfigure[3th-order tensor]{
    \begin{minipage}[t]{0.22\linewidth}
    \includegraphics[height = 2.5cm,width=3cm]{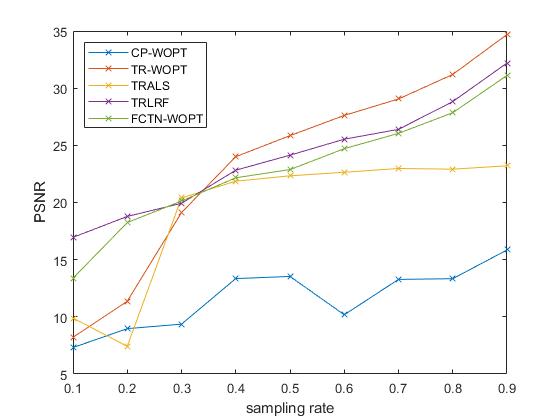}
    \end{minipage}
    }
    \subfigure[4th-order tensor]{
    \begin{minipage}[t]{0.22\linewidth}
    \includegraphics[height = 2.5cm,width=3cm]{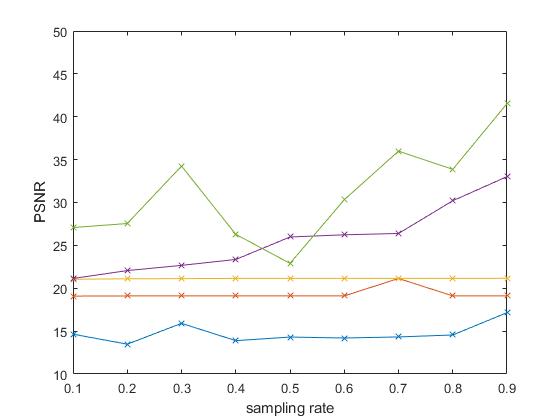}
    \end{minipage}    
    }
    \subfigure[5th-order tensor]{
    \begin{minipage}[t]{0.22\linewidth}
    \includegraphics[height = 2.5cm,width=3cm]{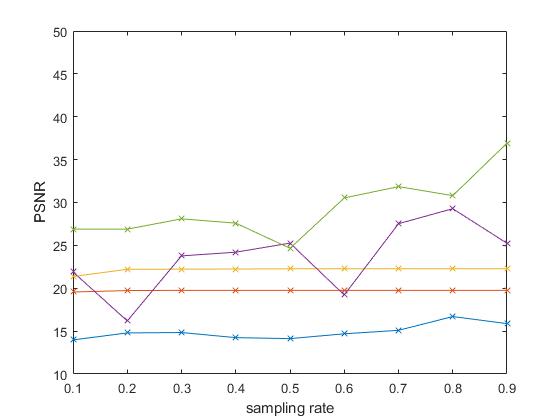}
    \end{minipage}    
    }
    \subfigure[6th-order tensor]{
    \begin{minipage}[t]{0.22\linewidth}
    \includegraphics[height = 2.5cm,width=3cm]{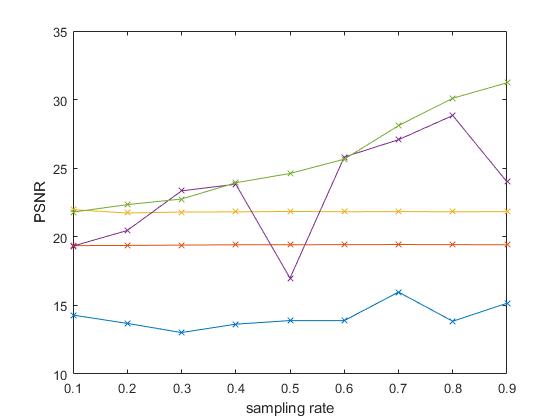}
    \end{minipage}    
    }
    \caption{Performance of different algorithms at different sampling rates}
    \centering
    \end{figure}

    It is easy to see that our proposed FCTN-WOPT algorithm has a significant advantage over various 
    algorithms except when the data being processed is of third order. This is because the proposed 
    algorithm is essentially a tensor ring (TR) decomposition algorithm when the original data is 
    of third order.

    \begin{table}[]
        \centering
        \caption{Performance of different algorithms with different order tensors}
        \setlength{\tabcolsep}{0.8mm}{
        \begin{tabular}{cc|c|c|c|c|c}
        \hline
        \hline
        \multicolumn{2}{c|}{}                     & CP-WOPT & TR-WOPT & TR-ALS & TRLRF & FCTN-WOPT{\ul } \\ \hline
        \multicolumn{1}{c|}{\multirow{2}{*}{3D}} & PSNR & 9.2312 & 18.3352 & 16.6218 & $\bf{18.7775}$ & 18.7091       \\ \cline{2-7} 
        \multicolumn{1}{c|}{}                  & SSIM & 0.0977 & 0.4540 & 0.4507 & 0.5010 & $\bf{0.5061}$       \\ \hline
        \multicolumn{1}{c|}{\multirow{2}{*}{4D}} & PSNR & 13.4781 & 19.1045 & 21.1007 & 22.0609 & $\bf{27.5505}$       \\ \cline{2-7} 
        \multicolumn{1}{c|}{}                  & SSIM & 0.1147 & 0.4592 & 0.511 & 0.5868 & $\bf{0.7027}$       \\ \hline
        \multicolumn{1}{c|}{\multirow{2}{*}{5D}} & PSNR & 14.7899 & 19.7286 & 22.2120 & 16.1910 & $\bf{26.8900}$       \\ \cline{2-7} 
        \multicolumn{1}{c|}{}                  & SSIM & 0.0479 & 0.4888 & 0.5674 & 0.4278 & $\bf{0.7186}$       \\ \hline
        \multicolumn{1}{c|}{\multirow{2}{*}{6D}} & PSNR & 13.684 & 19.3695 & 21.7351 & 20.4593 & $\bf{22.3457}$       \\ \cline{2-7} 
        \multicolumn{1}{c|}{}                  & SSIM & 0.1194 & 0.5006 & 0.5809 & 0.5303 & $\bf{0.5960}$       \\ \hline
        \hline
        \end{tabular}}
        \label{tab1}
        \centering
        \end{table}
    
    \subsection{Real data experiments}
    This experiment was mainly tested using real data. To better demonstrate the advantages of higher order data 
    processing, we chose some data from images and videos for the experiment. The algorithms used, we still use 
    those mentioned above for comparison, and their rank and termination conditions are all determined with the 
    same settings. 
    
    $\bf{Image~data~completion.}$
    We tested and compared the effect of different algorithms to complete the hyperspectral image with a size 
    of 60$\times$60$\times$20$\times$20, at a sampling rate of 0.2 The completion outcome is shown in Fig.3 with 
    the residuals.
    
    \begin{figure}[t]
        \centering
        \subfigure[Sampling]{
        \begin{minipage}[t]{0.2\linewidth}
        \includegraphics[height = 2.5cm,width=2.5cm]{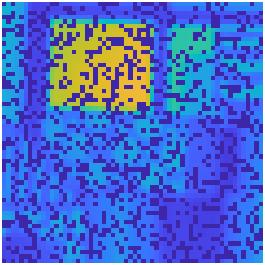}
        \end{minipage}    
        }
        \subfigure[CP-WOPT]{
        \begin{minipage}[t]{0.2\linewidth}
        \includegraphics[height = 2.5cm,width=2.5cm]{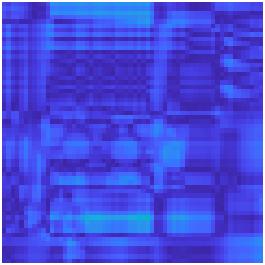}
        \end{minipage}    
        }
        \subfigure[TR-WOPT]{
        \begin{minipage}[t]{0.2\linewidth}
        \includegraphics[height = 2.5cm,width=2.5cm]{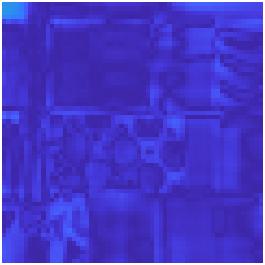}
        \end{minipage}    
        }
    
        \subfigure[TRALS]{
        \begin{minipage}[t]{0.2\linewidth}
        \includegraphics[height = 2.5cm,width=2.5cm]{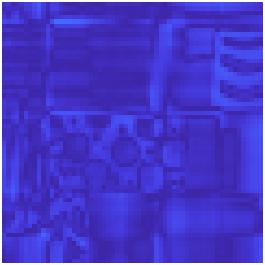}
        \end{minipage}    
        }
        \subfigure[TRLRF]{
        \begin{minipage}[t]{0.2\linewidth}
        \includegraphics[height = 2.5cm,width=2.5cm]{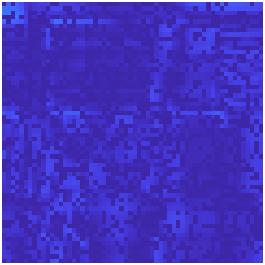}
        \end{minipage}    
        }
        \subfigure[$\bf{proposed}$]{
        \begin{minipage}[t]{0.2\linewidth}
        \includegraphics[height = 2.5cm,width=2.5cm]{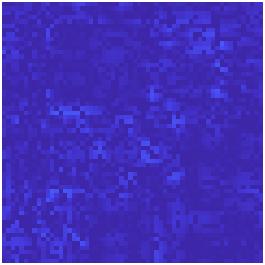}
        \end{minipage}    
        }
    \caption{The performance of different algorithms is compared for a sampling rate of 0.3. We 
    selected the HSV hyperspectral image dataset, size 60 × 60 × 20 × 20. We greyed out all hyperspectral 
    images and produced residual images of the hyperspectral images. In addition, image pixels in yellow indicate a large residual value, blue indicates a small residual value.}
    \end{figure}   
    
    From the experimental results we can clearly observe that the hue of the residual images is 
    more blue and the performance is more optimised than that of other algorithms. This is because 
    the rearrangement of the tensor modes shifts the correlation between them, leading to its 
    superiority over the traditional decomposition method.
    
    $\bf{Video~Data~Completion.}$
    We test two videos, the first one is HSV with a size of 144$\times$176$\times$3$\times$50 (width $\times$ height $\times$ 
    number of colour channels $\times$ number of video frames) and the second one is callphone with a size 
    of 144$\times$176$\times$3$\times$382 (width $\times$ height $\times$ number of colour channels $\times$ number of video frames).
    We also test under the same termination conditions and with the same tensor rank. 
    The number of iterations is set to 2000, $tol = 10^{-5}$, and the rank of the factor 
    is chosen to be 3. The results of the program are shown in Figure 4.
    
    In these video processing results, we obtain experimental results 
    that are more outstanding, also due to the compressibility of the decomposition algorithm itself, with the same 
    tensor rank but representing more information content.
    
    \begin{figure*}
    
        \centering
        \subfigure{
        \begin{minipage}{0.11\linewidth}
        \includegraphics[height = 1.5cm,width=1.5cm]{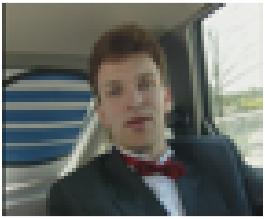}
        \end{minipage}
        }
        \subfigure{
        \begin{minipage}{0.11\linewidth}
        \includegraphics[height = 1.5cm,width=1.5cm]{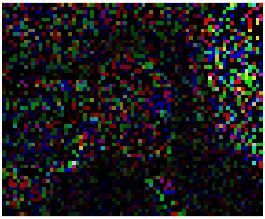}
        \end{minipage}    
        }
        \subfigure{
        \begin{minipage}{0.11\linewidth}
        \includegraphics[height = 1.5cm,width=1.5cm]{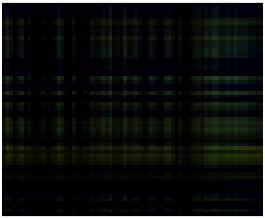}
        \end{minipage}    
        }
        \subfigure{
        \begin{minipage}{0.11\linewidth}
        \includegraphics[height = 1.5cm,width=1.5cm]{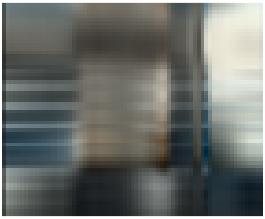}
        \end{minipage}    
        }
        \subfigure{
        \begin{minipage}{0.11\linewidth}
        \includegraphics[height = 1.5cm,width=1.5cm]{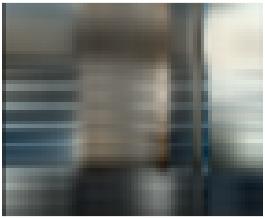}
        \end{minipage}    
        }
        \subfigure{
        \begin{minipage}{0.11\linewidth}
        \includegraphics[height = 1.5cm,width=1.5cm]{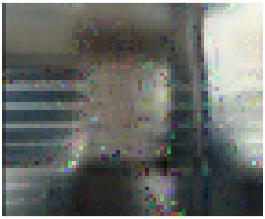}
        \end{minipage}    
        }
        \subfigure{
        \begin{minipage}{0.11\linewidth}
        \includegraphics[height = 1.5cm,width=1.5cm]{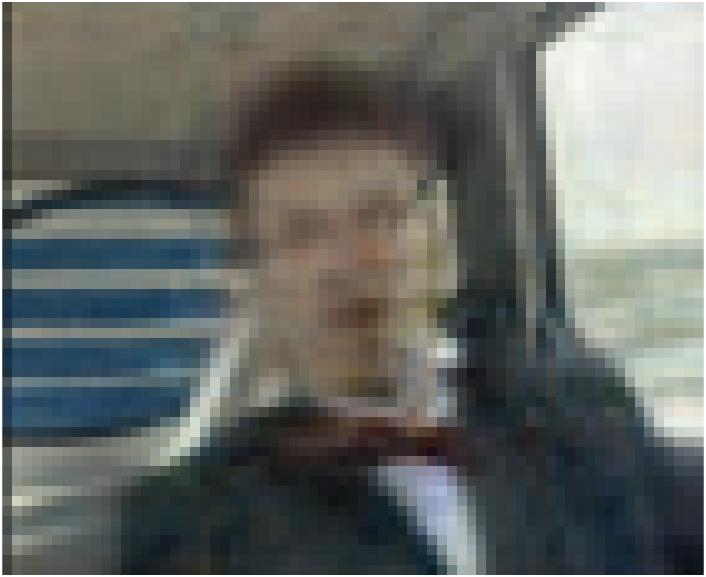}
        \end{minipage}    
        }
    
        \centering
        \subfigure{
        \begin{minipage}{0.11\linewidth}
        \includegraphics[height = 1.5cm,width=1.5cm]{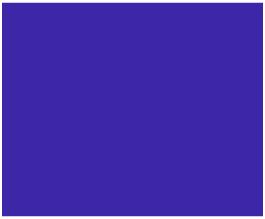}
        \end{minipage}
        }
        \subfigure{
        \begin{minipage}{0.11\linewidth}
        \includegraphics[height = 1.5cm,width=1.5cm]{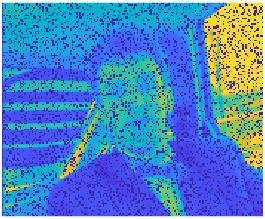}
        \end{minipage}    
        }
        \subfigure{
        \begin{minipage}{0.11\linewidth}
        \includegraphics[height = 1.5cm,width=1.5cm]{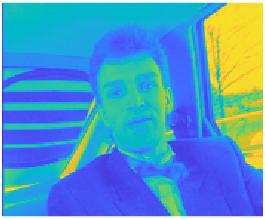}
        \end{minipage}    
        }
        \subfigure{
        \begin{minipage}{0.11\linewidth}
        \includegraphics[height = 1.5cm,width=1.5cm]{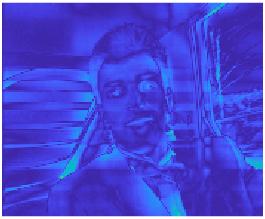}
        \end{minipage}    
        }
        \subfigure{
        \begin{minipage}{0.11\linewidth}
        \includegraphics[height = 1.5cm,width=1.5cm]{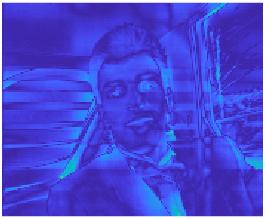}
        \end{minipage}    
        }
        \subfigure{
        \begin{minipage}{0.11\linewidth}
        \includegraphics[height = 1.5cm,width=1.5cm]{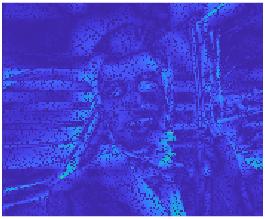}
        \end{minipage}    
        }
        \subfigure{
        \begin{minipage}{0.11\linewidth}
        \includegraphics[height = 1.5cm,width=1.5cm]{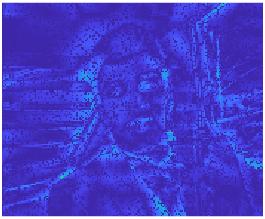}
        \end{minipage}    
        }
    
        \centering
        \subfigure{
        \begin{minipage}{0.11\linewidth}
        \includegraphics[height = 1.5cm,width=1.5cm]{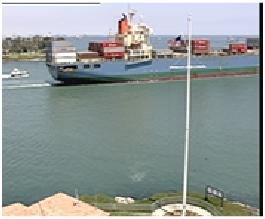}
        \end{minipage}
        }
        \subfigure{
        \begin{minipage}{0.11\linewidth}
        \includegraphics[height = 1.5cm,width=1.5cm]{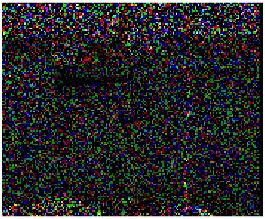}
        \end{minipage}    
        }
        \subfigure{
        \begin{minipage}{0.11\linewidth}
        \includegraphics[height = 1.5cm,width=1.5cm]{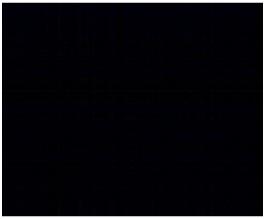}
        \end{minipage}    
        }
        \subfigure{
        \begin{minipage}{0.11\linewidth}
        \includegraphics[height = 1.5cm,width=1.5cm]{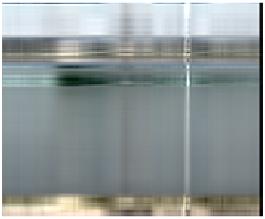}
        \end{minipage}    
        }
        \subfigure{
        \begin{minipage}{0.11\linewidth}
        \includegraphics[height = 1.5cm,width=1.5cm]{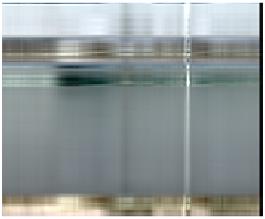}
        \end{minipage}    
        }
        \subfigure{
        \begin{minipage}{0.11\linewidth}
        \includegraphics[height = 1.5cm,width=1.5cm]{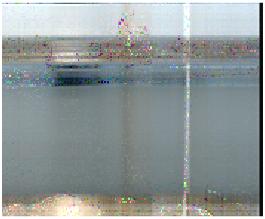}
        \end{minipage}    
        }
        \subfigure{
        \begin{minipage}{0.11\linewidth}
        \includegraphics[height = 1.5cm,width=1.5cm]{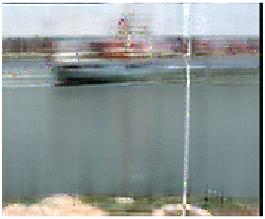}
        \end{minipage}    
        }
    
        \centering
        \subfigure{
        \begin{minipage}{0.11\linewidth}
        \includegraphics[height = 1.5cm,width=1.5cm]{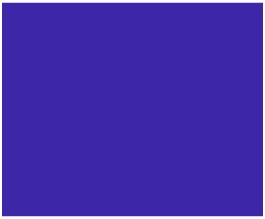}
        \end{minipage}
        }
        \subfigure{
        \begin{minipage}{0.11\linewidth}
        \includegraphics[height = 1.5cm,width=1.5cm]{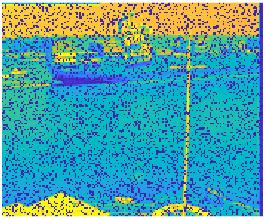}
        \end{minipage}    
        }
        \subfigure{
        \begin{minipage}{0.11\linewidth}
        \includegraphics[height = 1.5cm,width=1.5cm]{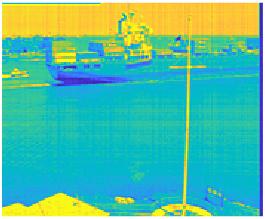}
        \end{minipage}    
        }
        \subfigure{
        \begin{minipage}{0.11\linewidth}
        \includegraphics[height = 1.5cm,width=1.5cm]{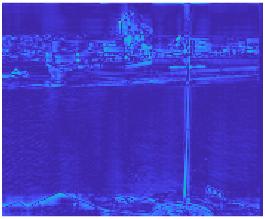}
        \end{minipage}    
        }
        \subfigure{
        \begin{minipage}{0.11\linewidth}
        \includegraphics[height = 1.5cm,width=1.5cm]{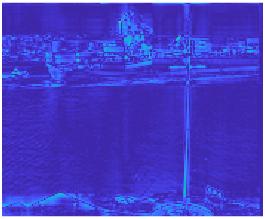}
        \end{minipage}    
        }
        \subfigure{
        \begin{minipage}{0.11\linewidth}
        \includegraphics[height = 1.5cm,width=1.5cm]{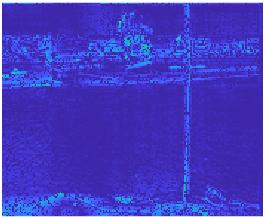}
        \end{minipage}    
        }
        \subfigure{
        \begin{minipage}{0.11\linewidth}
        \includegraphics[height = 1.5cm,width=1.5cm]{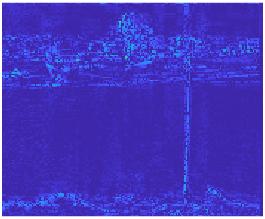}
        \end{minipage}    
        }
        \footnotesize\flushleft{~~~~~~Original~~~~Sampling~~~CP-OPT~~TR-WOPT~~~TRALS~~~~~TRLRF~~~~$\bf{proposed}$}
        \caption{The performance of the different algorithms is compared for a sampling rate 
        of 0.2. The first row is the completion of each method after sampling the callphone 
        part of the video, and the second row of data represents the residual image, which we 
        obtained by subtracting the first frame of its original video from the first frame of 
        the reconstructed video and taking the absolute value. The third row is the ship video 
        data, and the fourth row is its residual image.}
    
    \end{figure*}    
    
    \begin{figure*}
    
        \centering
        \subfigure{
        \begin{minipage}{0.14\linewidth}
        \includegraphics[height = 1.6cm,width=1.8cm]{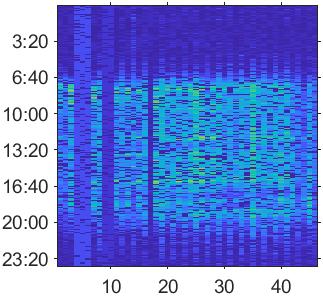}
        \end{minipage}    
        }
        \subfigure{
        \begin{minipage}{0.14\linewidth}
        \includegraphics[height = 1.6cm,width=1.8cm]{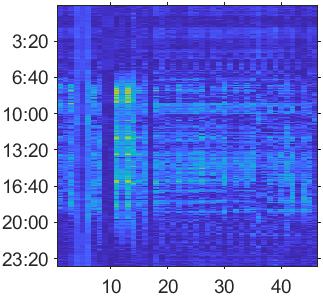}
        \end{minipage}    
        }
        \subfigure{
        \begin{minipage}{0.14\linewidth}
        \includegraphics[height = 1.6cm,width=1.8cm]{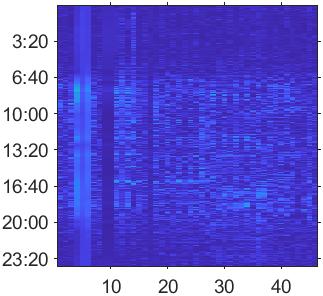}
        \end{minipage}    
        }
        \subfigure{
        \begin{minipage}{0.14\linewidth}
        \includegraphics[height = 1.6cm,width=1.8cm]{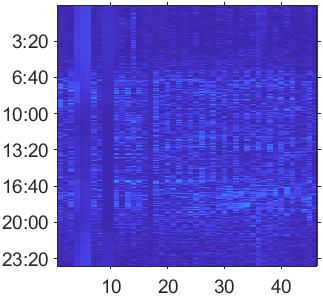}
        \end{minipage}    
        }
        \subfigure{
        \begin{minipage}{0.14\linewidth}
        \includegraphics[height = 1.6cm,width=1.8cm]{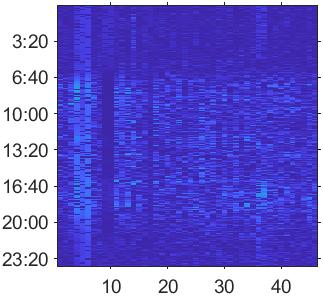}
        \end{minipage}    
        }
        \subfigure{
        \begin{minipage}{0.14\linewidth}
        \includegraphics[height = 1.6cm,width=1.8cm]{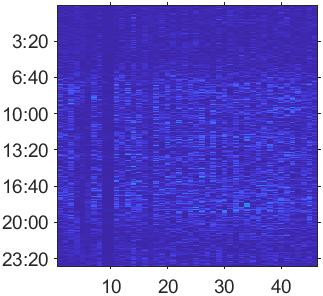}
        \end{minipage}    
        }
    
        \centering
        \subfigure{
        \begin{minipage}{0.14\linewidth}
        \includegraphics[height = 1.6cm,width=1.8cm]{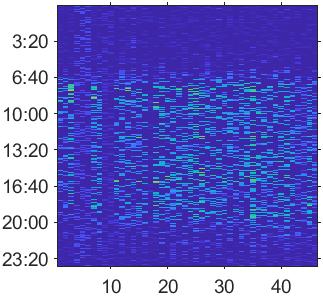}
        \end{minipage}    
        }
        \subfigure{
            \begin{minipage}{0.14\linewidth}
                \includegraphics[height = 1.6cm,width=1.8cm]{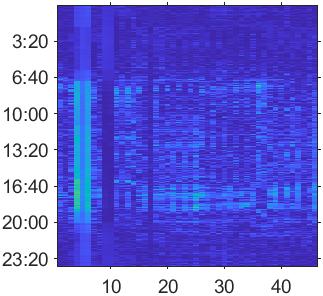}
        \end{minipage}    
        }
        \subfigure{
            \begin{minipage}{0.14\linewidth}
                \includegraphics[height = 1.6cm,width=1.8cm]{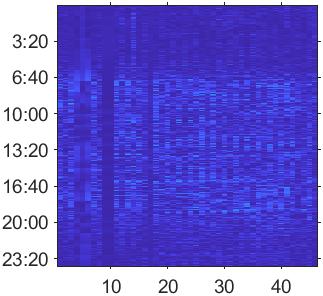}
        \end{minipage}    
        }
        \subfigure{
            \begin{minipage}{0.14\linewidth}
                \includegraphics[height = 1.6cm,width=1.8cm]{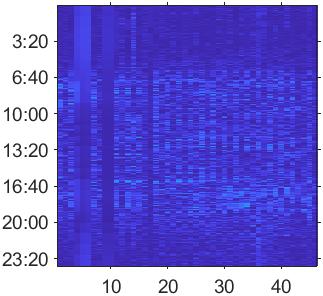}
        \end{minipage}    
        }
        \subfigure{
            \begin{minipage}{0.14\linewidth}
                \includegraphics[height = 1.6cm,width=1.8cm]{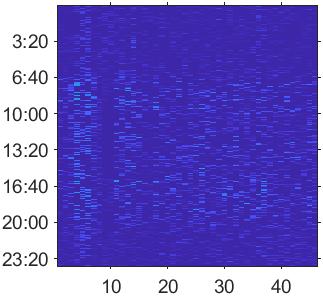}
        \end{minipage}    
        }
        \subfigure{
            \begin{minipage}{0.14\linewidth}
                \includegraphics[height = 1.6cm,width=1.8cm]{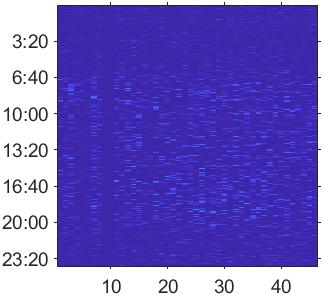}
        \end{minipage}    
        }
        \footnotesize\flushleft{~~~~Sampling~~~~~~~CP-OPT~~~~~~TR-WOPT~~~~~~TRALS~~~~~~~~TRLRF~~~~~~~$\bf{proposed}$}
        \caption{These experimental results represent a residual plot of the corresponding results from 
        the first day. The sampling rate of the images in the first row is 0.2 and the sampling rate of 
        the images in the second row is 0.7.}
    
    \end{figure*}
    
    $\bf{Traffic~Data~Completion.}$ This experiment uses the traffic flow dataset provided by 
    Grenoble Traffic Lab, which collects traffic flow data from 46 different road segments 
    over a period of 244 days, and measures every 15 seconds. We selected the 30 days of data 
    for the experiment and summed them to obtain a new dataset of 30-day, minute-by-minute 
    measurements for 49 road segments. The size of the dataset is 60$\times$24$\times$30$\times$46 (minutes $\times$ 
    hours $\times$ days $\times$ number of road segments). Then we sampled some data randomly and conducted tensor 
    completion experiments. The sampling rate for the first group was 0.2 and the sampling rate for the 
    second group was 0.7. The design of the rank of these tensor was the same, both of size 3. Their 
    maximum number of iterations was also the same, of size 1000. the minimum error was set to 10-4. 
    the experimental results are shown in Figure 5We selected the middle 30 days of data for the experiment 
    and summed them to obtain a new dataset of 30-day, minute-by-minute measurements for 49 road segments.  
    The size of the dataset is 60$\times$24$\times$30$\times$46 (minutes $\times$ hours $\times$ days $\times$ number of road segments). Then 
    we sampled some data randomly and conducted tensor completion experiments.The sampling rate for the first 
    set was 0.2 and the sampling rate for the second set was 0.7. The design of the rank of these 
    tensor completions is the same, all of size 3. Their maximum number of iterations is also the same, 
    number 1000. the minimum error is set to $10^{-4}$. the experimental results are shown in Figure 5.
    
    We can observe in Fig. 5 that the algorithm proposed in this paper also has a good tensor completion 
    result on the traffic flow data. This also validates the superior performance of the proposed algorithm 
    on high-dimensional tensor.

    \section{CONCLUSIONS}
    This algorithm proposes a Fully-Connected Tensor Network weighted optimization algorithm, which 
    is better than advanced algorithms for completion and is able to recover the original data with 
    a small amount of data. And our proposed algorithm, using the lbfgs gradient descent method, speeds 
    up the convergence of the algorithm's descent speed. In particular, tensor completion works better 
    for tensors of order four and above. However, the algorithm in this paper usually has to choose 
    different ranks for testing in order to get the desired results, which is time and effort consuming. 
    It is also not very useful in removing disturbances when external noise is present. Our next work 
    will be to create a tensor completion algorithm that automatically finds the optimal rank and 
    resists disturbances.


\begin{thebibliography}{8}


\bibitem{b1} Kolda T G, Bader B W. Tensor decompositions and applications[J]. SIAM review, 2009, 51(3): 455-500..
\bibitem{b2} Oseledets I V. Tensor-train decomposition[J]. SIAM Journal on Scientific Computing, 2011, 33(5): 2295-2317.
\bibitem{b3} Zhao Q, Zhou G, Xie S, et al. Tensor ring decomposition[J]. arXiv preprint arXiv:1606.05535, 2016.
\bibitem{b4} Q. Song, H. Ge, J. Caverlee, and X. Hu, “Tensor completion algorithmsin big data analytics,” ACM Trans. Knowl. Discovery Data, vol. 13, pp.1–48, 2019.
\bibitem{b5} Bazerque J A, Mateos G, Giannakis G B. Rank regularization and Bayesian inference for tensor completion and extrapolation[J]. IEEE transactions on signal processing, 2013, 61(22): 5689-5703.
\bibitem{b6} Ding, M.; Huang, T.-Z.; Ji, T.-Y.; Ji, T.-Y.; Zhao, X.-L.; and Yang,J.-H. 2019. Low-Rank Tensor Completion Using Matrix Factorization Based on Tensor Train Rank and Total Variation. Journalof Scientific Computing 81(2): 941–964.
\bibitem{b7} Gandy S, Recht B, Yamada I. Tensor completion and low-n-rank tensor recovery via convex optimization[J]. Inverse problems, 2011, 27(2): 025010.
\bibitem{b8} Yu D, Deng L, Seide F. The deep tensor neural network with applications to large vocabulary speech recognition[J]. IEEE Transactions on audio, speech, and language processing, 2012, 21(2): 388-396.
\bibitem{b9} Mahyari A G, Zoltowski D M, Bernat E M, et al. A tensor decomposition-based approach for detecting dynamic network states from EEG[J]. IEEE Transactions on Biomedical Engineering, 2016, 64(1): 225-237.
\bibitem{b10} Guo X, Huang X, Zhang L, et al. Support tensor machines for classification of hyperspectral remote sensing imagery[J]. IEEE Transactions on Geoscience and Remote Sensing, 2016, 54(6): 3248-3264.
\bibitem{b11} Zheng Y B, Huang T Z, Zhao X L, et al. Fully-Connected Tensor Network Decomposition and Its Application to Higher-Order Tensor Completion[C]//Proceedings of the AAAI Conference on Artificial Intelligence. 2021, 35(12): 11071-11078.
\bibitem{b12} Hu W, Tao D, Zhang W, et al. The twist tensor nuclear norm for video completion[J]. IEEE transactions on neural networks and learning systems, 2016, 28(12): 2961-2973.
\bibitem{b13} Yuan M, Zhang C H. On tensor completion via nuclear norm minimization[J]. Foundations of Computational Mathematics, 2016, 16(4): 1031-1068.
\bibitem{b14} Yu J, Li C, Zhao Q, et al. Tensor-ring nuclear norm minimization and application for visual: Data completion[C]//ICASSP 2019-2019 IEEE international conference on acoustics, speech and signal processing (ICASSP). IEEE, 2019: 3142-3146.
\bibitem{b15} Yuan L, Cao J, Zhao X, et al. Higher-dimension tensor completion via low-rank tensor ring decomposition[C]//2018 Asia-Pacific Signal and Information Processing Association Annual Summit and Conference (APSIPA ASC). IEEE, 2018: 1071-1076.
\bibitem{b16} Liu Y Y, Zhao X L, Song G J, et al. Fully-Connected Tensor Network Decomposition for Robust Tensor Completion Problem[J]. arXiv preprint arXiv:2110.08754, 2021.
\bibitem{b17} Ahad A, Long Z, Zhu C, et al. Hierarchical tensor ring completion[J]. arXiv preprint arXiv:2004.11720, 2020.
\bibitem{b18} Evrim Acar, Daniel M Dunlavy, Tamara G Kolda, and Morten Mørup.Scalable tensor factorizations for incomplete data. Chemometrics andIntelligent Laboratory Systems, 106(1):41–56, 2011.
\bibitem{b19} Yuan L, Li C, Mandic D, et al. Tensor ring decomposition with rank minimization on latent space: An efficient approach for tensor completion[C]//Proceedings of the AAAI Conference on Artificial Intelligence. 2019, 33(01): 9151-9158.
\bibitem{b20} Wang W, Aggarwal V, Aeron S. Efficient low rank tensor ring completion[C]//Proceedings of the IEEE International Conference on Computer Vision. 2017: 5697-5705.
\bibitem{b21} Chen Y, He W, Yokoya N, et al. Nonlocal tensor-ring decomposition for hyperspectral image denoising[J]. IEEE Transactions on Geoscience and Remote Sensing, 2019, 58(2): 1348-1362.
\bibitem{b22} He W, Chen Y, Yokoya N, et al. Hyperspectral super-resolution via coupled tensor ring factorization[J]. Pattern Recognition, 2022, 122: 108280.



\end{thebibliography}
\end{document}